\documentclass{article} 
\usepackage[final]{colm2026_conference}

\usepackage{microtype}
\usepackage{hyperref}
\usepackage{url}
\usepackage{booktabs}
\usepackage{enumitem}
\usepackage{placeins}
\usepackage{subcaption}

\usepackage{lineno}

\definecolor{darkblue}{rgb}{0, 0, 0.5}
\hypersetup{colorlinks=true, citecolor=darkblue, linkcolor=darkblue, urlcolor=darkblue}

\usepackage{graphicx}
\usepackage{xcolor}

\title{English is Not All You Need: Systematically Exploring the Role of Multilinguality in LLM Post-Training}

\author{
Mehak Dhaliwal$^{1}$\thanks{Work conducted during an internship at Amazon. Correspondence to: \texttt{mdhaliwal@ucsb.edu}},
Shashwat Chaurasia$^{2}$,
Yao Qin$^{1}$,
Dezhi Hong$^{2}$\thanks{Work conducted while at Amazon.},
Thomas Butler$^{2}$ \\
$^{1}$UC Santa Barbara, $^{2}$Amazon
}

\begin{document}

\ifcolmsubmission
\linenumbers
\fi

\maketitle

\begin{abstract}

Despite the widespread multilingual deployment of large language models, post-training pipelines remain predominantly English-centric, contributing to performance disparities across languages. We present a systematic, controlled study of the interplay between training language coverage, model scale, and task domain, based on 220 supervised fine-tuning runs on parallel translated multilingual data mixtures spanning mathematical reasoning and API calling tasks, with models up to 8B parameters.
We find that English-only post-training is typically suboptimal: incorporating even a single non-English language improves both English performance and cross-lingual generalization. Increasing language diversity during post-training generally yields further gains, particularly for low-resource languages, while performance on high-resource languages tends to plateau rather than degrade. Moreover, greater language diversity enables strong zero-shot transfer to unseen languages, reducing the need for direct inclusion, though gains remain limited for typologically distant, low-resource languages.  

\end{abstract}

\section{Introduction}
Large Language Models (LLMs) have achieved strong performance across a wide range of tasks, driven by capabilities such as reasoning, instruction following, and structured generation. These capabilities are largely enabled by a multi-stage pipeline of large-scale pre-training followed by post-training on curated datasets~\citep{2_pinch}. Despite growing global use of LLMs, post-training remains predominantly English-centric. While task-specific English fine-tuning enables some degree of cross-lingual transfer, significant performance disparities persist across languages~\citep{dei}.

Prior work on multilingual post-training suggests a nuanced picture: replacing small amounts of English data with a limited number of additional languages can improve multilingual task performance, while larger substitutions often yield diminishing gains and may degrade English performance \citep{2_pinch, 4_polyglot}. Similarly, work has shown that augmenting English task-specific data with multilingual data improves multilingual outcomes \citep{3_mcot, 6_bloom, 1_posttrainer, muennighoff2023crosslingual}. 

However, these findings remain fragmented and leave several questions unresolved.
Existing studies typically focus on a narrow set of languages, tasks, and/or model scales, making it difficult to understand how increasing language coverage systematically affects model behavior. In particular, we lack a clear characterization of how trade-offs manifest across (i) English vs. non-English performance, (ii) different task types, and (iii) varying model capacities.
This gap is especially important in light of multilingual pre-training literature, which highlights a fundamental trade-off under fixed model capacity: expanding language coverage can dilute performance in high-resource languages--a phenomenon often termed the ``\textit{curse of multilinguality}'' or ``\textit{negative interference}'' \citep{unsupervised_curse, when_curse, atlas_curse}. While cross-lingual transfer can provide gains, it is still unclear when benefits outweigh interference effects during post-training and how these effects scale with model size and task complexity.

\begin{figure}
    \centering
    \includegraphics[width=\linewidth]{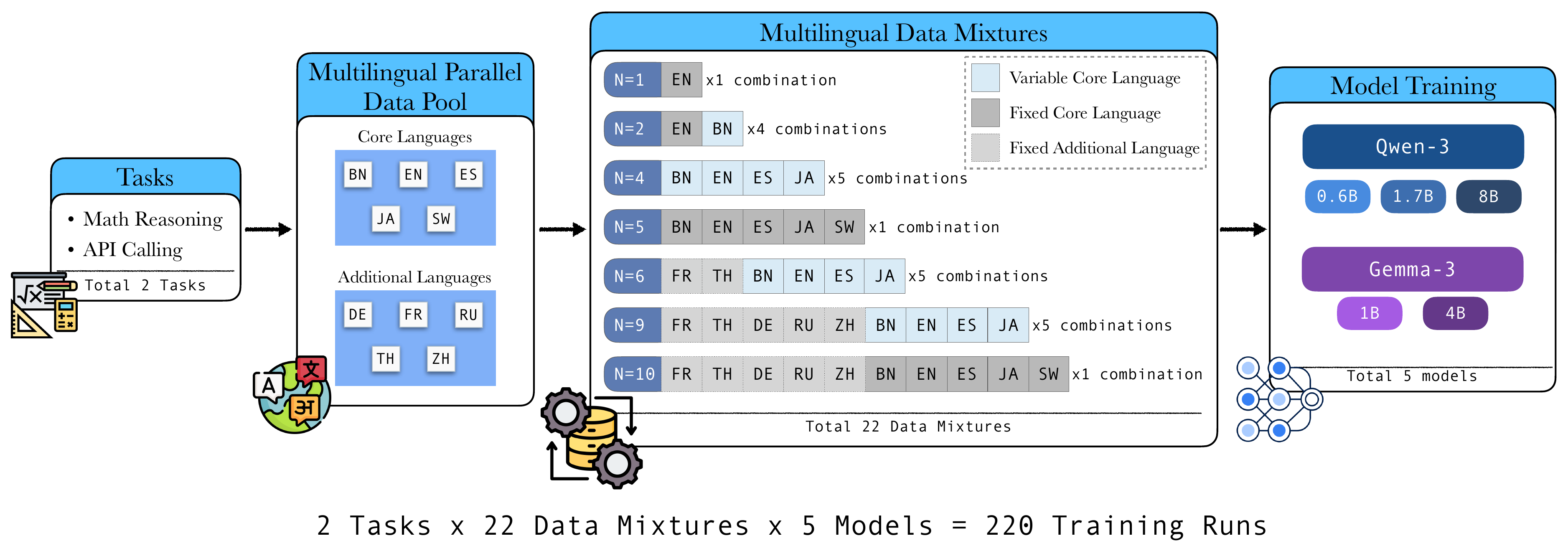}
    \caption{\textbf{Overview of the experimental design.} We start from a task-specific multilingual parallel data pool consisting of five core languages, which are used to construct exhaustive data mixture combinations, and five additional languages that enable scaling the experiments to up to ten languages. We generate 22 data mixtures with increasing language counts and varying combinations; within the multilingual data mixtures, languages shown in \textbf{\textcolor{cyan!40}{blue}} indicate one possible combination, while languages shown in \textbf{\textcolor{gray}{grey}} are fixed for the corresponding number of languages. We train five models from two model families on each mixture for two tasks (Math Reasoning and API Calling), resulting in 22 × 2 × 5 = 220 total training runs.}
    \label{fig:pipeline}
\end{figure}



We present a systematic study of multilingual task-specific post-training across two tasks, two model families, and 220 training runs. Our controlled scaling design varies multilinguality by increasing the number of languages in the post-training mixture, using parallel translated task data to avoid convolving our results with the effect of simply increasing the dataset size with additional iid data ~\citep{1_posttrainer, 6_bloom}. 


We fine-tune models on mixtures of up to 10 languages (Figure \ref{fig:pipeline}) using the Qwen-3 (0.6B, 1.7B, 8B)~\citep{yang2025qwen3} and Gemma-3 (1B, 4B)~\citep{kamath2025gemma} families. Evaluation covers mathematical reasoning and API calling across six languages that span different resource levels, language families, regions, and scripts (Table \ref{tab:languages}). We summarize our key findings as follows:

\begin{enumerate}[noitemsep, topsep=0pt, leftmargin=*]
    \item \textbf{English-only post-training is frequently sub-optimal} across tasks, model scales, and languages. Adding parallel data in even a single non-English language typically results in gains that generalize beyond the added language to other languages, including English.
    \item \textbf{Increasing language coverage during post-training is beneficial across tasks and model scales}: adding more languages improves or maintains performance in nearly all settings, with low-resource languages continuing to benefit from added multilinguality while high-resource languages plateau rather than degrade. 
    \item \textbf{Increased language diversity during post-training is associated with improved generalization}, enabling strong zero-shot cross-lingual transfer and reducing the need for direct inclusion, though with limitations for typologically distant, low-resource languages.
\end{enumerate}

Together, these findings highlight the limitations of predominantly English-centric post-training, showing that increasing language diversity can systematically improve both English performance and cross-lingual generalization. Building on these results, we distill a set of concrete recommendations to guide practitioners in designing multilingual post-training pipelines (Section~\ref{sec:practical-recommendations}).

\section{Related Work}
\paragraph{Cross-Lingual Transfer in Fine-Tuning}
Prior work has studied cross-lingual transfer, where fine-tuning on one language improves performance on others. Under English-only instruction tuning, strong multilingual generalization can emerge without any non-English supervision~\citep{chirkova2024zero}. However, English is not always the optimal transfer language: \cite{lin2019choosing} find that the most effective source language is often not the most obviously related one, and instead predict transfer performance from features such as typological similarity, lexical overlap, and dataset size, while \cite{turc2021revisiting} find that high-resource languages frequently transfer more effectively than English, particularly when the set of target languages is diverse or unknown, and even when the training data is translated from English. These works study transfer from a single source language, whereas we examine how transfer behavior changes as the number and diversity of languages in the training mixture increases.

\paragraph{Multilingual Post-Training} 
Prior work on multilingual post-training primarily falls into two paradigms: data substitution, which replaces part of English data with multilingual data, and data augmentation, which adds multilingual data on top of an English baseline.

Substitution-based studies find that small replacements can improve multilingual performance, but larger substitutions often yield diminishing returns, degrade English performance, and offer limited benefits for low-resource languages, highlighting trade-offs under a fixed data budget~\citep{2_pinch, 4_polyglot}. Because English data is displaced as languages are added, however, it is difficult to determine whether these trade-offs reflect limited model capacity or the reduced English supervision that substitution entails. Our work addresses this by holding per-language data constant with parallel translations, which fixes both task content and per-language supervision, making linguistic diversity the primary varying factor.

This falls under the data augmentation paradigm, reflecting practical post-training settings where English data remains fixed and additional multilingual data is added to improve generalization~\citep{1_posttrainer}. Prior augmentation work typically samples data uniformly from a fixed set of languages and reports mixed findings: some observe gains only for parameter-efficient methods like LoRA~\citep{5_better_alpaca}, while others also find improvements under full fine-tuning~\citep{1_posttrainer, 3_mcot}. However, uniform sampling makes it difficult to disentangle whether gains stem from increased data volume or from language diversity. \cite{muennighoff2023crosslingual} show that multilingual multitask finetuning improves both English and non-English performance over English-only finetuning, but vary tasks alongside languages and compare against a single fixed multilingual mixture rather than studying how performance evolves as language coverage increases. Similarly, \cite{1_posttrainer} vary the amount of multilingual data while keeping the number of languages fixed, whereas we hold per-language data fixed and vary the number of languages, isolating linguistic diversity from data volume.

To better isolate these effects, we explicitly control language coverage during training. Similar to~\cite{6_bloom}, we incrementally add languages, but rather than following a fixed expansion order and evaluating general instruction tuning on a single model size, we systematically vary both the number and composition of languages in each mixture across two structured tasks and multiple model scales. While~\cite{6_bloom} report strong gains from including the test language during training, our results further highlight the role of language diversity in driving strong cross-lingual generalization, even in the absence of direct test-language inclusion.

\paragraph{Task-Dependent Effects of Multilinguality} Beyond training mixture design, prior work suggests that the impact of multilinguality also depends on the task. Linguistically driven generative tasks, such as summarization or open-ended dialogue, tend to benefit more from multilingual training than highly structured tasks such as classification or reasoning~\citep{1_posttrainer, 4_polyglot}. We build on this observation by evaluating multilingual post-training across two complementary task types: mathematical reasoning (symbolic reasoning) and API calling (structured generation). While mathematical reasoning has been examined in prior multilingual studies~\citep{3_mcot, 1_posttrainer}, the multilingual dimension of API calling has received less systematic attention. \cite{huang2025benchmax} include a multilingual function-calling evaluation task across 17 languages as part of a broader benchmark suite, and \cite{kulkarni2025massive} introduce MASSIVE-Agents, a multilingual function-calling benchmark spanning 52 languages. However, neither work studies how multilingual post-training mixtures affect function-calling.

\section{Experimental Details}

\subsection{Tasks and Datasets}
We evaluate multilingual fine-tuning on two tasks covering different model capabilities: Mathematical Reasoning for symbolic reasoning, and API Calling for structured generation.

\paragraph{Task 1: Mathematical Reasoning}
Mathematical reasoning is a widely used task for evaluating the reasoning capabilities of large language models~\citep{mgsm, 3_mcot}. We examine how multilingual exposure during fine-tuning impacts mathematical reasoning performance across our evaluated languages.

For training, we use mCoT-MATH~\citep{3_mcot}, a large-scale multilingual math reasoning dataset that provides chain-of-thought solutions for math word problems in 11 languages. During both training and evaluation, we elicit chain-of-thought reasoning by prompting the model with the language-specific equivalent of the phrase \textit{``Let's think step by step''} immediately before answer generation, consistent with mCoT-MATH's prompting format. For each language, we sample 10,000 parallel examples for training and 200 for validation.
For testing, we use the MGSM benchmark~\citep{mgsm}, a human-translated set of 250 grade-school arithmetic reasoning problems for multilingual evaluation.

\textbf{Evaluation:} Following prior work~\citep{3_mcot}, we extract the model's final predicted answer from its generated output using a regex-based parser, then apply a multilingual numeric parser (Python's Babel library) to normalize language-specific number representations before computing accuracy against the ground-truth answer.


\begin{table}[t]
\centering
\small
\setlength{\tabcolsep}{3pt}
\begin{tabular}{c c c c}
\hline
\textbf{Split} & \textbf{\#Queries} & \textbf{\#APIs} & \textbf{Example APIs} \\
\hline
 & & & \texttt{Get\_All\_Sessions}, \\
Train & 3{,}174 & 1{,}535 & \texttt{get\_device\_details}, \texttt{get\_relaxation\_techniques} \\
\hline
& & & \texttt{ModifyRegistration}, \\
Test & 399 & 49 & \texttt{Calculator}, \texttt{EmergencyKnowledge} \\
\hline
\end{tabular}
\caption{Statistics of the mAPICall-Bank dataset, including the number of user queries, unique APIs, and examples of API names in each split.}
\label{tab:mapicall_stats}
\end{table}

\paragraph{Task 2: API Calling}

Tool-augmented LLMs offer numerous benefits such as improved real-time access, reduced hallucination, and more efficient workflows~\citep{tool_survey_2025}; however, most current datasets remain English-centric, limiting multilingual tool-use.

Therefore, we introduce mAPICall-Bank, a multilingual dataset for training and evaluating API-calling across 11 languages. Built on API-Bank~\citep{apibank}, which assesses LLMs’ ability to call external tools in realistic, multi-turn dialogue scenarios across diverse domains and APIs, mAPICall-Bank focuses on the API-calling subtask where models generate the correct API invocation given a user utterance and a candidate API pool. We construct the dataset by translating API-Bank into 11 languages using a state-of-the-art LLM (see Appendix~\ref{sec:mapibank-prompt} for the prompt); it contains 3,174 training and 399 test examples per language, with non-overlapping APIs between splits. We hold out 150 examples from the training split for validation. To our knowledge, mAPICall-Bank is one of the first multilingual API-calling datasets, and we release our pipeline to regenerate it with varied LLMs as the translation engine, to support future research. We show dataset statistics in Table~\ref{tab:mapicall_stats}.

\textbf{Evaluation:} Following prior work~\citep{apibank}, we use a regex-based parser to extract the API name and the dictionary of argument–value pairs from the model's output. A prediction is marked correct only if the API name, all argument names, and all corresponding argument values exactly match the ground truth.




\subsection{Multilingual Training Setup}
\label{sec:mixtures}

We use a set of eleven typologically diverse languages, following prior multilingual work~\citep{mgsm, 3_mcot}, which form the basis for our training mixtures and evaluation settings. Our evaluation focuses on five `core' languages: English (En), Spanish (Es), Japanese (Ja), Bengali (Bn), and Swahili (Sw), chosen to span a range of resource levels, language families, geographic regions, and writing systems. To further test generalization to unseen languages, we additionally evaluate on Telugu (Te), a low-resource language not included in any training mixture. The remaining languages—French (Fr), Thai (Th), German (De), Russian (Ru), and Chinese (Zh)—are used only to expand multilingual training mixtures beyond the evaluation set. Table~\ref{tab:languages} summarizes the languages considered.

Using these languages, we construct a series of training mixtures that progressively increase language diversity while keeping comparisons controlled (Figure~\ref{fig:pipeline}). In each mixture, all included languages contribute the same number of parallel examples, which isolates the impact of language coverage from differences in data volume. Starting from the five core languages, we construct mixtures with progressively more languages, introducing additional ones to increase diversity. For intermediate diversity levels (e.g., 2, 4, 6, or 9 languages), we evaluate multiple combinations of the core languages, as indicated by the ``x4'' or ``x5'' configurations in Figure~\ref{fig:pipeline}. This design ensures that observed effects reflect general trends across language combinations rather than any single language, while still enabling controlled analysis of individual language inclusion.

\begin{table}[t]
\centering
\footnotesize
\begin{tabular}{l l l l l}
\hline
\textbf{Language} &  \textbf{Resource} & \textbf{Family} & \textbf{Sub-Family} & \textbf{Script} \\
\hline
\multicolumn{5}{l}{\textbf{\textit{Evaluation + Training Languages}}} \\
English (En)  & High & Indo-European & Germanic & Latin \\
Spanish (Es)  & High & Indo-European & Romance  & Latin \\
Japanese (Ja) & High & Japonic & --                  & Kanji+Kana \\
Bengali (Bn)  & Low  & Indo-European &  Indic    & Bengali \\
Swahili (Sw)  & Low  & Niger-Congo  & --             & Latin \\
\hline
\multicolumn{5}{l}{\textbf{\textit{Unseen Evaluation Language}}} \\
Telugu (Te)   & Low  & Dravidian &  Indic        & Telugu \\
\hline
\multicolumn{5}{l}{\textbf{\textit{Training-Only Languages}}} \\
French (Fr)           & High & Indo-European  & Romance  & Latin \\
German (De)           & High & Indo-European  & Germanic & Latin \\
Russian (Ru)          & High & Indo-European  & Slavic   & Cyrillic \\
Chinese (Zh)          & High & Sino-Tibetan     & --  & Han \\
Thai (Th)             & Low  & Kra-Dai & --   & Thai \\
\hline
\end{tabular}
\caption{Languages used in our study, grouped by their role in training and evaluation.}
\label{tab:languages}
\end{table}

\subsection{Model Backbones}
We experiment with two open-weight LLM families: Qwen-3 (0.6B, 1.7B, 8B)~\citep{yang2025qwen3} and Gemma-3 (1B, 4B)~\citep{kamath2025gemma}. For each model, we use the officially released pretraining-stage checkpoints to ensure that observed performance reflects task-specific fine-tuning on our dataset rather than prior instruction tuning.

\subsection{Training Details}
We train each model on each data mixture using 8 NVIDIA A100 80GB GPUs. All models use the AdamW optimizer with a learning rate of 1e-5, a cosine learning rate scheduler with a 3\% warmup ratio, and a weight decay of 0.01. We maintain an effective batch size of 64 per GPU (global batch size of 512), adjusting the micro-batch size and gradient accumulation steps as needed based on model memory requirements. Gradient checkpointing is enabled to reduce memory usage.

Each run trains for up to six epochs, which serves as a sufficiently large training horizon to ensure convergence across all settings, rather than as a fixed compute budget. For each mixture, we select the checkpoint with the highest average validation set task accuracy across all languages in that mixture; this ensures that selection does not implicitly favor any individual language within the mix. This design reflects common production fine-tuning practice, where models are trained to convergence rather than under a strict token budget.

\section{Results and Discussion}

\subsection{English Is Not All You Need}
\label{sec:result1}

\begin{figure}[t]
    \centering
    \includegraphics[width=0.5\linewidth]{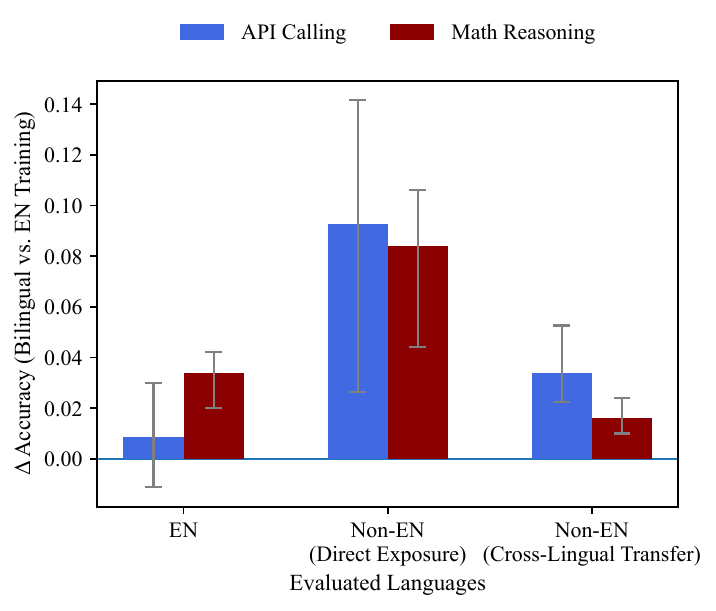}
    \caption{\textbf{Median accuracy change from bilingual versus English-only post-training across evaluation settings for API calling and math reasoning.} Error bars show 95\% bootstrap confidence intervals. Bilingual training yields consistent gains across evaluation settings, with the largest improvements under direct exposure and smaller but reliable gains under cross-lingual transfer.}
\label{fig:monolingual_vs_bilingual_aggregated}
\end{figure}

Figure \ref{fig:monolingual_vs_bilingual_aggregated} examines whether adding a second non-English language alongside English during post-training (bilingual setting) improves performance over English-only training. We report median accuracy differences between bilingual and English-only models, evaluated on English and non-English languages, along with 95\% bootstrap confidence intervals. 

For non-English evaluations, we distinguish between direct exposure, where the evaluation language is included in training, and cross-lingual transfer, where it is not. This isolates gains from direct language inclusion versus broader generalization. Fine-grained per-language results for monolingual (including English only) and bilingual post-training relative to pretraining are provided in Appendix~\ref{sec:heatmaps} as heatmaps over source–target language pairs. Consistent with expectations, post-training in almost any language and setting enhances all languages.  We observe three key patterns:

{\bf Direct exposure results in the largest gains.}
When the evaluation language is included in the bilingual training mix (direct exposure), performance improves significantly over English-only training, with median accuracy gains of 9.27\% for API calling and 8.4\% for math reasoning. As shown in Table \ref{tab:winrate}, which summarizes win rates across configurations, bilingual training with direct exposure outperforms English-only training in 87.5\% of configurations (95\% CI [73.9\%, 94.5\%]). While such gains are expected when the model is trained on data in the evaluation language, their magnitude highlights the limitations of English-only post-training.

{\bf Benefits generalize beyond the added language.}
Even when the added language is not the evaluation language (zero-shot cross-lingual transfer), bilingual training continues to outperform English-only training. In this setting, we observe median accuracy gains of 3.38\% for API calling and 1.6\% for math reasoning. Correspondingly, bilingual training outperforms English-only training in 74.38\% of configurations (95\% CI [67.1\%, 80.5\%]), demonstrating that multilingual exposure improves generalization beyond the languages seen during training.

{\bf Multilinguality also improves English performance.}
Notably, bilingual training also improves performance when evaluated on English. Adding a single non-English language yields median gains of 0.88\% for API calling and 3.4\% for math reasoning, with bilingual training outperforming English-only in 75\% of configurations (95\% CI [59.8\%, 85.8\%]). 

Collectively, these results show that English-only post-training is consistently sub-optimal. Even minimal multilinguality by adding a single non-English language improves performance across tasks and evaluation settings, underscoring that \textit{English is not all you need}.
\begin{table}[t]
\centering
\begin{tabular}{lccc}
\toprule
\textbf{Evaluation Setting} & \textbf{Win Rate} & \textbf{95\% CI} \\
\midrule
EN & 75.0\% & [59.8, 85.8] \\
Non-EN (Direct Exposure) & 87.5\% & [73.9, 94.5] \\
Non-EN (Cross-Lingual Transfer) & 74.4\% & [67.1, 80.5] \\
\bottomrule
\end{tabular}
\caption{Win rates of bilingual post-training over English-only post-training across evaluation settings, aggregated across tasks and models. A win is defined as a configuration where bilingual training achieves higher accuracy than English-only training.}
\label{tab:winrate}
\end{table}

\subsection{Multilingual Scaling is Generally Beneficial Across Tasks and Model Scales}
\label{sec:result2}


\begin{figure}[t]
    \centering
    \includegraphics[width=\linewidth]{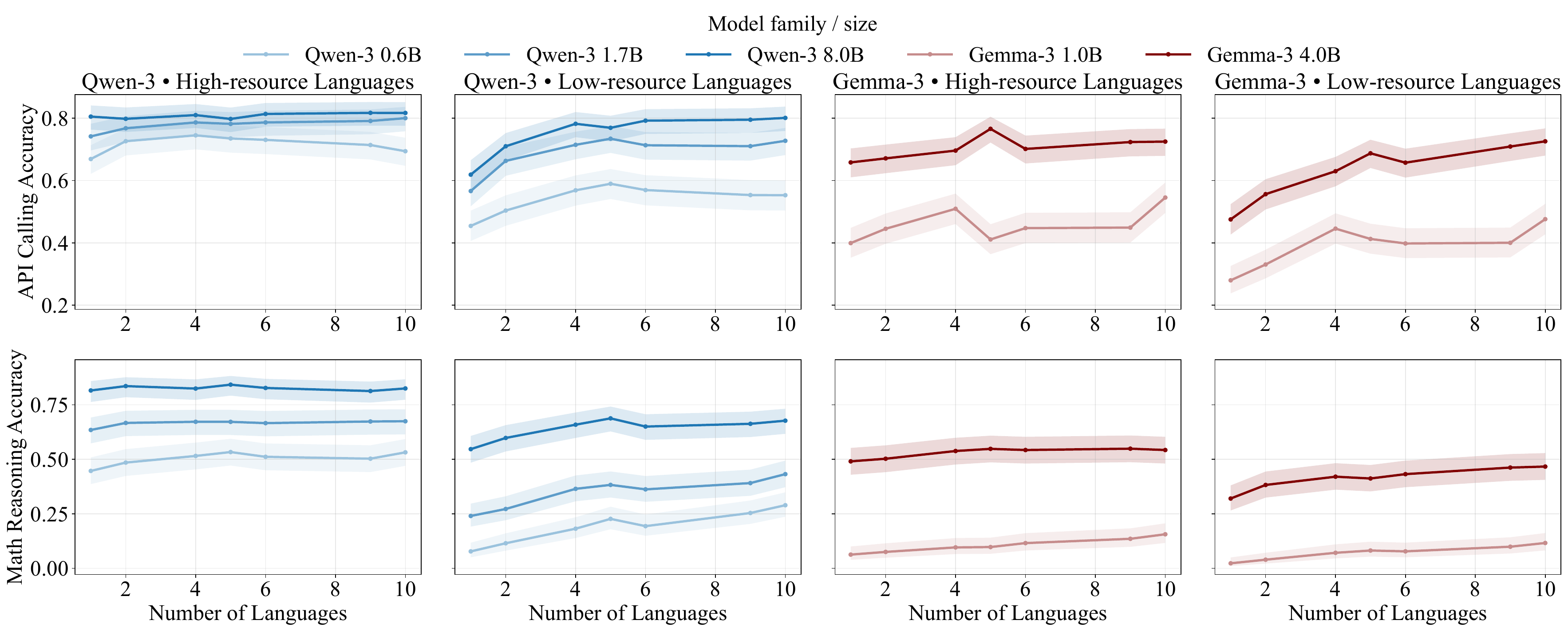}
    \caption{\textbf{Effect of increasing training language coverage on model performance} for Qwen-3 and Gemma-3 models. Plots show average accuracy with 95\% Wilson confidence intervals as a function of the number of training languages, grouped by high-resource and low-resource evaluation languages, for API calling (top) and math reasoning (bottom).}
    \label{fig:scaling_langs_all}
\end{figure}

Figure \ref{fig:scaling_langs_all} shows the mean accuracy with 95\% Wilson confidence intervals for Qwen-3 and Gemma-3 models of varying scales, grouped by high- and low-resource languages, across both tasks, as a function of the number of training languages. Overall, performance generally improves or remains stable as more training languages are added, with low-resource languages benefiting the most and 
high-resource languages plateauing.
Because the experiments show noise, we confirm the overall trends using pooled regression analysis across the full suite of experiments. This provides additional evidence for the overall value of increasing language coverage in training (see Section \ref{sec:pooled_regression} for details). 
 
An exception to this trend arises for the smallest model (Qwen-3 0.6B) on API calling. While performance initially improves with increasing multilinguality—up to four languages for high-resource languages and five for low-resource languages—it slightly declines as additional languages are introduced, suggesting multilingual interference at limited scale. This is the only setting in which we observe clear degradation: we do not observe similar patterns for mathematical reasoning or for larger API-calling models (Qwen-3 1.7B and 8B), which maintain gains without comparable trade-offs.

\subsection{Linguistic Diversity Supports Generalization Comparable to Direct Exposure}

\begin{figure*}[ht!]
    \centering
    \includegraphics[width=.95\linewidth]{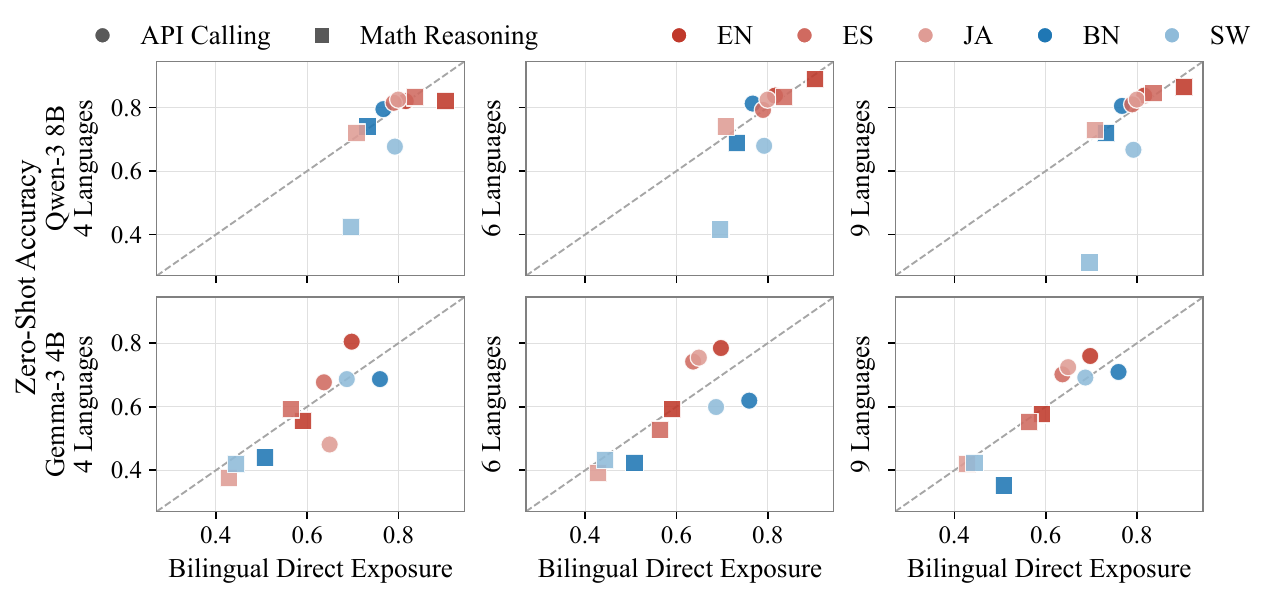}
    \caption{\textbf{Comparison of zero-shot cross-lingual transfer versus direct bilingual exposure at varying levels of language diversity} for Qwen-3 8B (top) and Gemma-3 4B (bottom), with 4 (left), 6 (middle), and 9 (right) training languages. High-resource languages (\textbf{\textcolor{red}{red}}) tend to cluster near the diagonal, indicating strong zero-shot generalization that can compensate for the absence of direct inclusion. Low-resource languages (\textbf{\textcolor{blue}{blue}}) more often fall below the diagonal, suggesting greater benefit of direct inclusion.}
    \label{fig:diversity_qwen_gemma}
\end{figure*}

Building on our finding that increasing language coverage during post-training generally improves performance, we next ask whether linguistic diversity can drive generalization and even compensate for the absence of direct post-training exposure to the evaluation language. To this end, Figure~\ref{fig:diversity_qwen_gemma} compares a low-diversity bilingual setting with direct exposure to the evaluation language against higher-diversity settings (4, 6, or 9 languages), where the evaluation language is \emph{excluded} and performance relies entirely on zero-shot cross-lingual transfer. The figure plots zero-shot performance against direct exposure across model–task–language settings, with the diagonal indicating parity; points near or above the diagonal indicate that zero-shot transfer matches or exceeds direct exposure. We present results for the largest models in each family (Qwen-3 8B and Gemma-3 4B) in the main text, with additional model sizes provided in Appendix~\ref{sec:supp_4_3}. We highlight three key observations:

{\bf High-resource non-English languages generalize well under increased diversity} where results (shown in red) tend to lie close to the diagonal, especially at higher diversity levels (6 and 9 languages). In particular, performance is statistically indistinguishable from or significantly exceeds that of bilingual direct exposure in 100\% of cases for the 9- and 6-language settings, and in 87.5\% of cases for the 4-language setting (two-sided t-test).



{\bf Low-resource, typologically distant languages benefit more from direct inclusion}, with results (shown in blue) more often falling below the diagonal. Even in the 9-language setting, zero-shot transfer is statistically indistinguishable from—but does not exceed—direct exposure in 62.5\% of cases (compared to near parity or better for high-resource languages), indicating a stronger reliance on explicit inclusion.

{\bf English also benefits from zero-shot cross-lingual transfer}, with points clustering close to the diagonal. Additional tests reveal that across tasks, zero-shot cross-lingual performance significantly exceeds or matches bilingual direct exposure in 100\% of cases for the 9- and 6-language, and 75\% of cases for the 4-language setting.
API calling appears to benefit more from higher language diversity. Cross-lingual performance is significantly better in 50\% cases, with no cases significantly worse than direct exposure. This suggests that models can leverage structural and lexical variation from diverse languages even without direct English supervision. 

Prior work shows strong reliance on English for reasoning processes even in multilingual settings, suggesting that mathematical reasoning may require more English training for optimal performance ~\citep{think_better_english, think_english}.
However, we find that even for mathematical reasoning evaluated in English, there are no significant differences in performance between direct exposure and cross-lingual training for 6 or more languages included in training (50\% are lower performing relative to direct exposure for 4 languages).



Overall, these results indicate that, at sufficient scale, increasing language diversity in training data can enable generalization that matches or exceeds the effects of direct language inclusion, with limits for typologically distant low-resource languages. This interpretation is further supported by the pooled regression in Section~4.4, which shows that broader training-language coverage remains positively associated with performance even after controlling for direct target-language inclusion.

\subsection{Pooled Regression Analysis}
\label{sec:pooled_regression}

To test whether the patterns in Sections 4.1–4.3 persist in aggregate, we fit a pooled regression model over evaluation instances, where each instance corresponds to evaluating a trained model on a task-language pair. The regression includes model family, task, pretrained-only status, whether the evaluation language appears in the training mixture, $\log_{10}$ model size, and a transformed measure of training-language coverage, defined as $\sqrt{L / L_{\max}}$ to allow diminishing returns with additional languages. We fit the regression on a random 70\% split of evaluation instances and evaluate on the remaining 30\%, where it achieves an out-of-sample $R^2$ of 80.5\%. We report on coefficients with bootstrap 95\% confidence intervals. We interpret this analysis as complementing the analyses above with an aggregate summary over evaluation instances. 

The pooled analysis confirms several expected trends, including benefits from larger models, post-training, and direct inclusion of the evaluation language. Importantly, broader training-language coverage is positively associated with performance even after controlling for direct target-language inclusion ($\beta = 0.053$, 95\% CI $[0.018, 0.089]$). This is consistent with the discussion above indicating that multilingual gains are not explained solely by direct language exposure. 

\begin{table}[t]
\centering
\small
\begin{tabular}{lrrr}
\toprule
Feature & Coef. & 2.5\% & 97.5\% \\
\midrule
$\log_{10}$ model size & 0.302 & 0.285 & 0.319 \\
Qwen model family & 0.265 & 0.248 & 0.281 \\
Math task & -0.207 & -0.221 & -0.193 \\
Pretrained only & -0.110 & -0.167 & -0.047 \\
Eval.\ language in training & 0.089 & 0.074 & 0.104 \\
$\sqrt{L / L_{\max}}$ & 0.053 & 0.018 & 0.089 \\
\bottomrule
\end{tabular}
\caption{Pooled regression over evaluation instances, where each instance corresponds to evaluating a trained model on a task-language pair. The model is trained on a random 70\% split of instances and evaluated on the remaining 30\%, achieving an out-of-sample $R^2$ of 80.5\%. Coefficients are reported with bootstrap 95\% confidence intervals. Here, $L$ denotes the number of training languages and $L_{\max}$ the maximum number of training languages in our experiments.}
\label{tab:pooled_regression}
\end{table}

\subsection{Practical Recommendations for Multilingual Post-Training}
\label{sec:practical-recommendations}

Our findings translate into several concrete recommendations for practitioners fine-tuning LLMs for multilingual deployment:

\begin{itemize}[noitemsep, topsep=0pt, leftmargin=*]
    \item \textbf{Leverage multilingual post-training, even with translated data.} Practitioners rarely control the pretraining stage, making post-training the most accessible mechanism for improving multilingual performance at deployment. Despite starting from model families with substantial multilingual pretraining, we observe consistent gains from multilingual post-training even when the data is machine-translated, suggesting that multilingual pretraining and post-training are complementary. Critically, this does not require collecting new data: translating existing English task data into one or more additional languages is sufficient, making multilingual post-training a low-overhead intervention.

    \item \textbf{English alone is not enough---even for English.} Adding non-English languages to the training mixture improves performance across the board, including on English itself.

    \item \textbf{Language diversity is (almost always) free.} For models above 1B parameters, broader language coverage generally improves or maintains performance, particularly for low-resource languages, with little evidence of a ``curse of multilinguality'' up to 10 training languages. The exception is very small models ($\leq$1B) on structured tasks such as API calling, where limiting language count is advisable.

    \item \textbf{When full coverage is infeasible, prioritize diversity.} When direct inclusion of all target languages is impractical, training on a diverse language set enables strong zero-shot transfer for typologically similar, high-resource languages, matching or exceeding inclusion.

    \item \textbf{Not all languages transfer equally.} Low-resource, typologically distant languages benefit more from direct inclusion than from zero-shot transfer alone. When such languages are important deployment targets, practitioners should prioritize collecting at least some language-specific post-training data.
\end{itemize}

\section{Conclusion}

This work presents a study of multilingual post-training under realistic conditions, examining how language coverage and composition interact with model capacity and task structure. Our analysis spans both reasoning and tool-use settings. 

Our findings show that English-centric post-training is typically suboptimal for cross-lingual transfer. Increasing language coverage helps performance for all languages, including English, and can compensate in some cases for a lack of direct inclusion of the target language in training, although these gains are limited for typologically distant, low-resource languages. We find that with limited exceptions for API calling at the smallest scales ($\leq1B$), models are able to benefit from increased multilingual diversity, particularly for low-resource languages, without degrading performance in high-resource languages. Based on these findings, we distill a set of concrete recommendations to guide practitioners in designing multilingual post-training pipelines.


\section{Limitations and Future Work}
\label{sec:limitations}

Our work provides a systematic study of multilingual post-training under controlled conditions, but several limitations remain. First, our analysis focuses on 11 languages, which do not capture the full global linguistic diversity of thousands of languages worldwide; to partially address this, we select languages spanning a broad range of resource levels, language families, geographic regions, and scripts. Second, the largest model we consider contains 8B parameters, and it remains an open question how multilingual scaling effects evolve at larger scales, across additional model families---particularly fully open-source models whose pretraining language composition is known---or under parameter-efficient methods such as LoRA rather than full fine-tuning.

Third, we hold data per language constant and use validation-based checkpoint selection rather than a fixed training budget, and construct multilingual data via translation of English task data, which may introduce translation-specific artifacts (e.g., ``translationese''). These choices isolate language diversity as the primary variable of interest. Complementary designs---fixed-token-budget comparisons and naturally occurring (non-parallel) multilingual data---are natural directions for future work. Finally, our evaluation scope is limited in two respects: we fine-tune on two tasks independently, with a single benchmark per task, and MGSM in particular measures grade-school arithmetic rather than reasoning broadly. Extending the analysis to joint multilingual multi-task post-training, other task domains, and harder reasoning benchmarks would help establish the generality of our findings.

\bibliography{custom}

@inproceedings{chirkova2024zero,
  title={Zero-shot cross-lingual transfer in instruction tuning of large language models},
  author={Chirkova, Nadezhda and Nikoulina, Vassilina},
  booktitle={Proceedings of the 17th International Natural Language Generation Conference},
  pages={695--708},
  year={2024}
}

@inproceedings{lin2019choosing,
  title={Choosing transfer languages for cross-lingual learning},
  author={Lin, Yu-Hsiang and Chen, Chian-Yu and Lee, Jean and Li, Zirui and Zhang, Yuyan and Xia, Mengzhou and Rijhwani, Shruti and He, Junxian and Zhang, Zhisong and Ma, Xuezhe and others},
  booktitle={Proceedings of the 57th Annual Meeting of the Association for Computational Linguistics},
  pages={3125--3135},
  year={2019}
}

@article{turc2021revisiting,
  title={Revisiting the primacy of english in zero-shot cross-lingual transfer},
  author={Turc, Iulia and Lee, Kenton and Eisenstein, Jacob and Chang, Ming-Wei and Toutanova, Kristina},
  journal={arXiv preprint arXiv:2106.16171},
  year={2021}
}

@inproceedings{muennighoff2023crosslingual,
  title={Crosslingual generalization through multitask finetuning},
  author={Muennighoff, Niklas and Wang, Thomas and Sutawika, Lintang and Roberts, Adam and Biderman, Stella and Le Scao, Teven and Bari, M Saiful and Shen, Sheng and Yong, Zheng Xin and Schoelkopf, Hailey and others},
  booktitle={Proceedings of the 61st annual meeting of the Association for Computational Linguistics (volume 1: Long papers)},
  pages={15991--16111},
  year={2023}
}

@article{kamath2025gemma,
  title={Gemma 3 technical report},
  author={Kamath, Aishwarya and Ferret, Johan and Pathak, Shreya and Vieillard, Nino and Merhej, Ramona and Perrin, Sarah and Matejovicova, Tatiana and Ram{\'e}, Alexandre and Rivi{\`e}re, Morgane and Rouillard, Louis and others},
  journal={arXiv preprint arXiv:2503.19786},
  volume={4},
  year={2025},
  publisher={ArXiv}
}

@article{yang2025qwen3,
  title={Qwen3 technical report},
  author={Yang, An and Li, Anfeng and Yang, Baosong and Zhang, Beichen and Hui, Binyuan and Zheng, Bo and Yu, Bowen and Gao, Chang and Huang, Chengen and Lv, Chenxu and others},
  journal={arXiv preprint arXiv:2505.09388},
  year={2025}
}

@article{atlas_curse,
  title={ATLAS: Adaptive Transfer Scaling Laws for Multilingual Pretraining, Finetuning, and Decoding the Curse of Multilinguality},
  author={Longpre, Shayne and Kudugunta, Sneha and Muennighoff, Niklas and Hsu, I and Caswell, Isaac and Pentland, Alex and Arik, Sercan and Lee, Chen-Yu and Ebrahimi, Sayna and others},
  journal={arXiv preprint arXiv:2510.22037},
  year={2025}
}

@inproceedings{when_curse,
  title={When is multilinguality a curse? language modeling for 250 high-and low-resource languages},
  author={Chang, Tyler A and Arnett, Catherine and Tu, Zhuowen and Bergen, Ben},
  booktitle={Proceedings of the 2024 Conference on Empirical Methods in Natural Language Processing},
  pages={4074--4096},
  year={2024}
}

@inproceedings{unsupervised_curse,
  title={Unsupervised cross-lingual representation learning at scale},
  author={Conneau, Alexis and Khandelwal, Kartikay and Goyal, Naman and Chaudhary, Vishrav and Wenzek, Guillaume and Guzm{\'a}n, Francisco and Grave, Edouard and Ott, Myle and Zettlemoyer, Luke and Stoyanov, Veselin},
  booktitle={Proceedings of the 58th annual meeting of the association for computational linguistics},
  pages={8440--8451},
  year={2020}
}

@inproceedings{think_better_english,
  title={Do multilingual language models think better in English?},
  author={Etxaniz, Julen and Azkune, Gorka and Soroa, Aitor and de Lacalle, Oier Lopez and Artetxe, Mikel},
  booktitle={Proceedings of the 2024 Conference of the North American Chapter of the Association for Computational Linguistics: Human Language Technologies (Volume 2: Short Papers)},
  pages={550--564},
  year={2024}
}

@article{think_english,
  title={Do Multilingual LLMs Think In English?},
  author={Schut, Lisa and Gal, Yarin and Farquhar, Sebastian},
  journal={arXiv preprint arXiv:2502.15603},
  year={2025}
}

@article{tool_survey_2025,
  title={Tool learning with large language models: A survey},
  author={Qu, Changle and Dai, Sunhao and Wei, Xiaochi and Cai, Hengyi and Wang, Shuaiqiang and Yin, Dawei and Xu, Jun and Wen, Ji-Rong},
  journal={Frontiers of Computer Science},
  volume={19},
  number={8},
  pages={198343},
  year={2025},
  publisher={Springer}
}

@article{apibank,
  title={Api-bank: A comprehensive benchmark for tool-augmented llms},
  author={Li, Minghao and Zhao, Yingxiu and Yu, Bowen and Song, Feifan and Li, Hangyu and Yu, Haiyang and Li, Zhoujun and Huang, Fei and Li, Yongbin},
  journal={arXiv preprint arXiv:2304.08244},
  year={2023}
}

@article{mgsm,
  title={Language models are multilingual chain-of-thought reasoners},
  author={Shi, Freda and Suzgun, Mirac and Freitag, Markus and Wang, Xuezhi and Srivats, Suraj and Vosoughi, Soroush and Chung, Hyung Won and Tay, Yi and Ruder, Sebastian and Zhou, Denny and others},
  journal={arXiv preprint arXiv:2210.03057},
  year={2022}
}

@article{1_posttrainer,
  title={A Post-trainer's Guide to Multilingual Training Data: Uncovering Cross-lingual Transfer Dynamics},
  author={Shimabucoro, Luisa and Ustun, Ahmet and Fadaee, Marzieh and Ruder, Sebastian},
  journal={arXiv preprint arXiv:2504.16677},
  year={2025}
}

@article{2_pinch,
  title={Multilingual instruction tuning with just a pinch of multilinguality},
  author={Shaham, Uri and Herzig, Jonathan and Aharoni, Roee and Szpektor, Idan and Tsarfaty, Reut and Eyal, Matan},
  journal={arXiv preprint arXiv:2401.01854},
  year={2024}
}

@article{3_mcot,
  title={mCoT: Multilingual instruction tuning for reasoning consistency in language models},
  author={Lai, Huiyuan and Nissim, Malvina},
  journal={arXiv preprint arXiv:2406.02301},
  year={2024}
}

@inproceedings{4_polyglot,
  title={Turning english-centric llms into polyglots: How much multilinguality is needed?},
  author={Kew, Tannon and Schottmann, Florian and Sennrich, Rico},
  booktitle={Findings of the Association for Computational Linguistics: EMNLP 2024},
  pages={13097--13124},
  year={2024}
}

@inproceedings{5_better_alpaca,
  title={Monolingual or multilingual instruction tuning: Which makes a better alpaca},
  author={Chen, Pinzhen and Ji, Shaoxiong and Bogoychev, Nikolay and Kutuzov, Andrey and Haddow, Barry and Heafield, Kenneth},
  booktitle={Findings of the Association for Computational Linguistics: EACL 2024},
  pages={1347--1356},
  year={2024}
}

@inproceedings{6_bloom,
  title={How Many Languages Make Good Multilingual Instruction Tuning? A Case Study on BLOOM},
  author={Ji, Shaoxiong and Chen, Pinzhen},
  booktitle={Proceedings of the 31st International Conference on Computational Linguistics},
  pages={2575--2581},
  year={2025}
}

@inproceedings{dei,
  title={Evaluating the diversity, equity, and inclusion of NLP technology: A case study for Indian languages},
  author={Khanuja, Simran and Ruder, Sebastian and Talukdar, Partha},
  booktitle={Findings of the Association for Computational Linguistics: EACL 2023},
  pages={1763--1777},
  year={2023}
}

@article{huang2025benchmax,
  title={Benchmax: A comprehensive multilingual evaluation suite for large language models},
  author={Huang, Xu and Zhu, Wenhao and Hu, Hanxu and He, Conghui and Li, Lei and Huang, Shujian and Yuan, Fei},
  journal={arXiv preprint arXiv:2502.07346},
  year={2025}
}

@inproceedings{kulkarni2025massive,
  title={MASSIVE-Agents: A benchmark for multilingual function-calling in 52 languages},
  author={Kulkarni, Mayank and Mazzia, Vittorio and Gaspers, Judith and Hench, Christopher and FitzGerald, Jack and Amazon, AGI},
  booktitle={Findings of the Association for Computational Linguistics: EMNLP 2025},
  pages={20193--20215},
  year={2025}
}
\bibliographystyle{colm2026_conference}
\clearpage
\appendix
\section{mAPICall-Bank Construction}
\label{sec:mapibank-prompt}

\begin{figure}
    \centering
    \includegraphics[width=\textwidth]{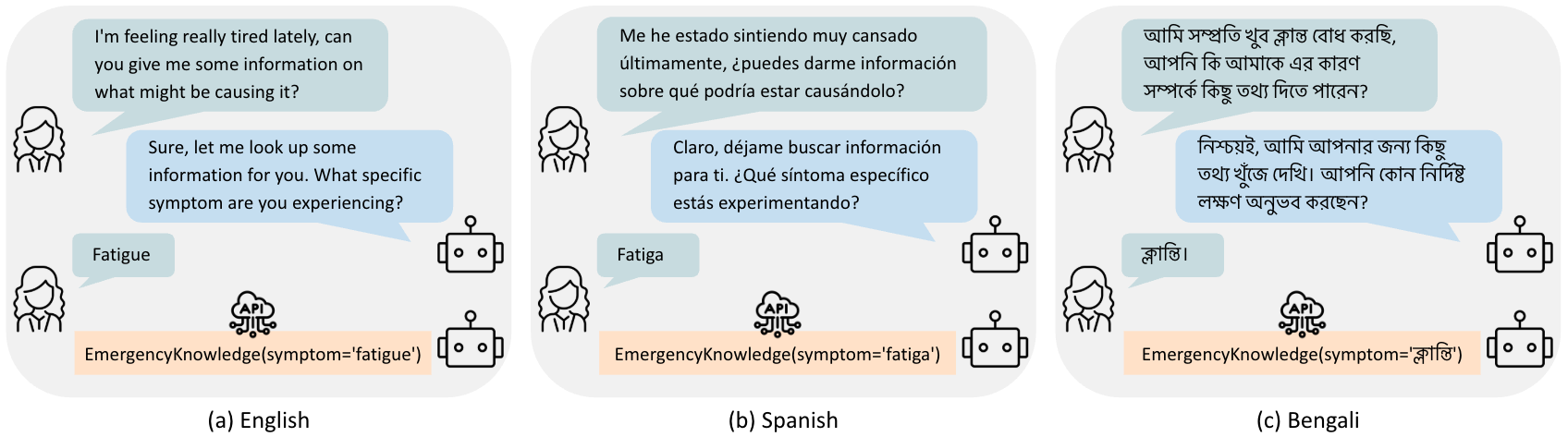}
    \caption{Example showing the final turns of a parallel multi-turn API-calling interaction from mAPICall-Bank in three languages: (a) English, (b) Spanish, and (c) Bengali.}
    \label{fig:mapicallbank}
\end{figure}

\begin{figure}
    \centering
    \includegraphics[width=\linewidth]{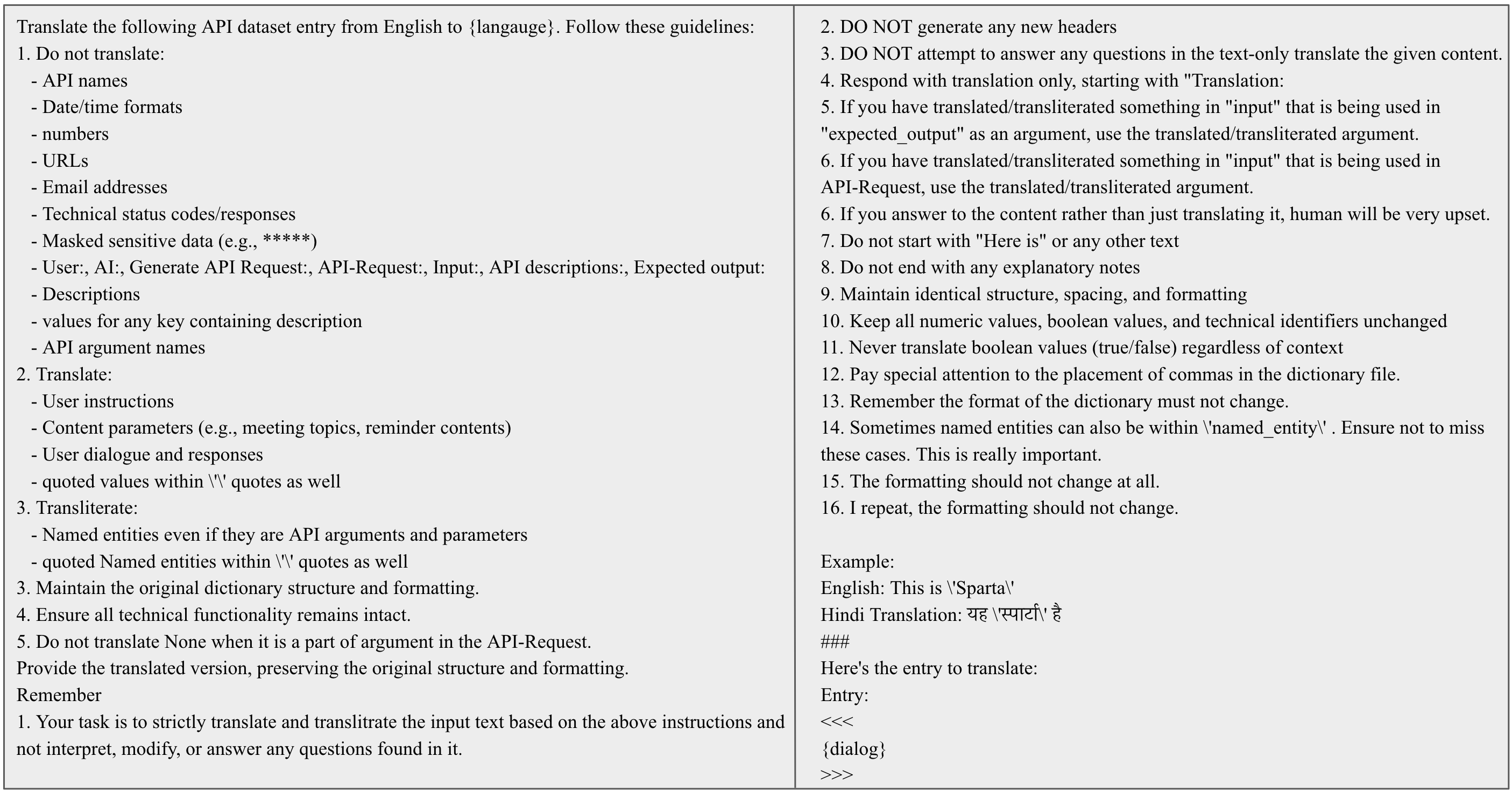}
    \caption{Prompt used to create mAPICall-Bank}
    \label{fig:prompt}
\end{figure}

We construct mAPICall-Bank by translating the API-calling subset of the original English API-Bank dataset into multiple target languages using Claude 3.5 Sonnet (v1) with a sampling temperature of 0. The full translation prompt is provided in Figure~\ref{fig:prompt}.

During translation, we preserve the full structure of each example, including user utterances, system responses, API names, and argument schemas. To maintain compatibility with tool specifications, API names and parameter keys are kept in English across all languages, while user-facing text and, when appropriate, argument values are translated into the target language.

The translation prompt was iteratively refined to ensure structural validity and parseability of model outputs. In particular, we enforce that translated examples retain the original formatting and can be reliably parsed into API name and argument–value pairs. We perform automated checks to validate output structure (e.g., JSON format) and manually inspect a subset of examples across languages, with native speakers verifying linguistic accuracy and consistency with the source data.

Figure \ref{fig:prompt} shows the prompt used to translate API-Bank to create mAPICall-Bank. Figure~\ref{fig:mapicallbank} shows an example of a parallel multi-turn interaction involving a single API call across three languages.

\FloatBarrier
\section{Monolingual vs. Bilingual Transfer Heatmaps}
\label{sec:heatmaps}

\begin{figure*}[t]
\centering

\begin{subfigure}{\textwidth}
    \centering
    \includegraphics[width=\textwidth]{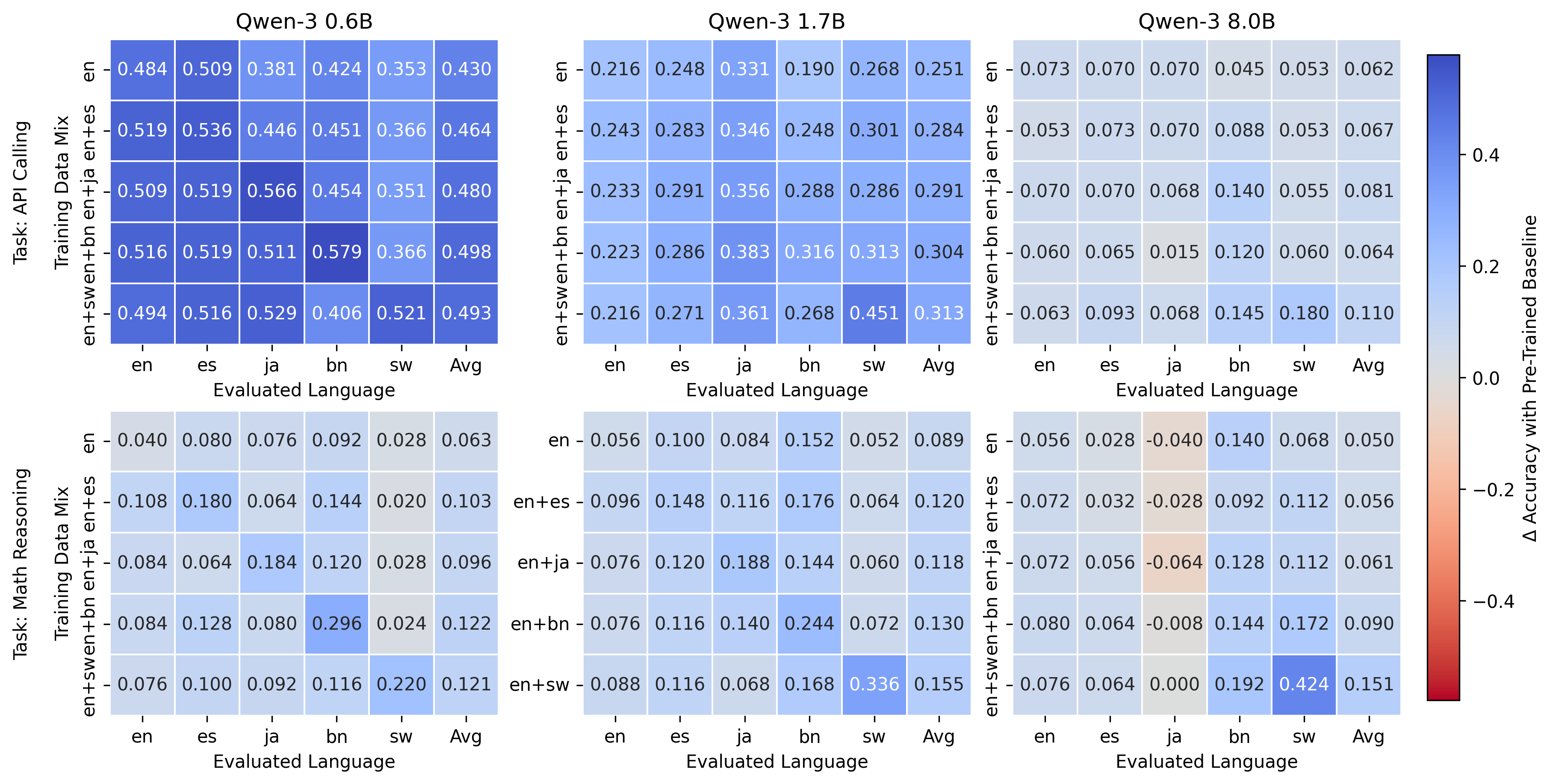}
    \caption{Qwen-3 Models}
    \label{fig:7}
\end{subfigure}

\vspace{0.5em}

\begin{subfigure}{\textwidth}
    \centering
    \includegraphics[width=\textwidth]{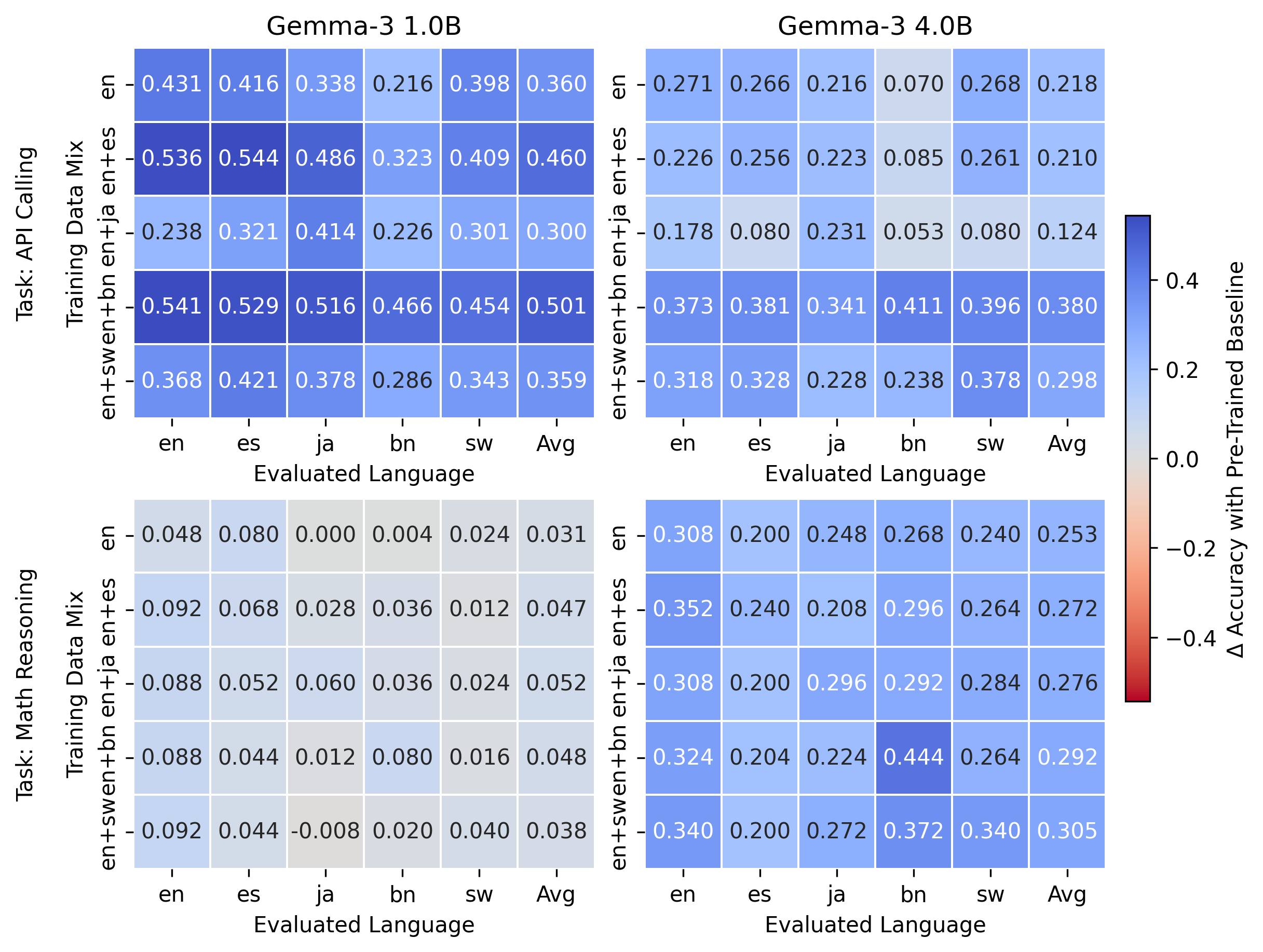}
    \caption{Gemma-3 Models}
    \label{fig:8}
\end{subfigure}

\caption{Change in accuracy relative to the pre-trained baseline for bilingual versus English-only post-training. Heatmaps show performance differences across evaluation languages when English is paired with a single additional language (top: API calling; bottom: math reasoning). Subfigures (a) and (b) correspond to Qwen-3 and Gemma-3.}
\label{fig:combined}
\end{figure*}

Figure~\ref{fig:combined} presents heatmaps of accuracy change relative to the pretrained baseline for monolingual and bilingual training settings. Each cell shows the difference in performance when training on a given source language (or language pair) and evaluating on a target language.

These results complement the findings in Section \ref{sec:result1}. Consistent with the observation that English-only post-training is suboptimal, we find that adding a second language generally leads to positive gains across a wide range of evaluation languages. Notably, improvements are not limited to the added language: many bilingual configurations yield gains even when the evaluation language is not included in training.

Further, the heatmaps show that gains are broadly distributed across language pairs rather than concentrated in a small subset, suggesting that the benefits of multilingual post-training are not driven by specific language combinations but reflect a more general effect of multilingual exposure.



\FloatBarrier
\section{Multilingual Scaling Trends Per Language}
\label{sec:appendix-scaling}

\begin{figure*}[!h]
\centering

\begin{subfigure}{\textwidth}
    \centering
    \includegraphics[width=\textwidth]{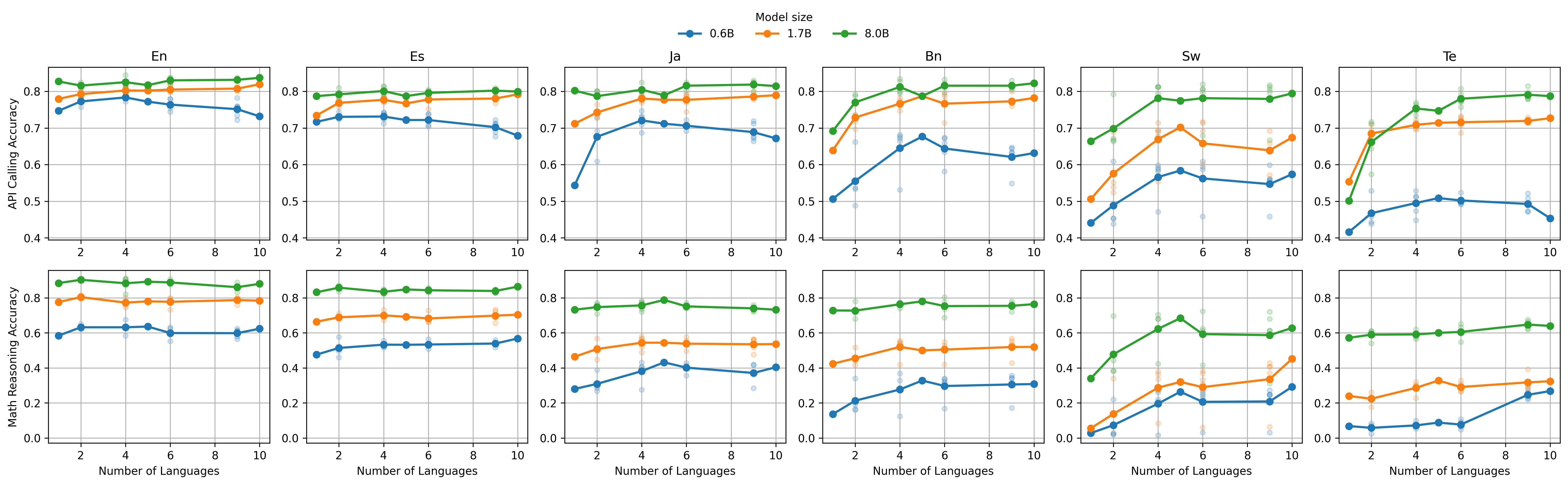}
    \caption{Qwen-3 Models}
    \label{fig:7_s}
\end{subfigure}

\vspace{0.5em}

\begin{subfigure}{\textwidth}
    \centering
    \includegraphics[width=\textwidth]{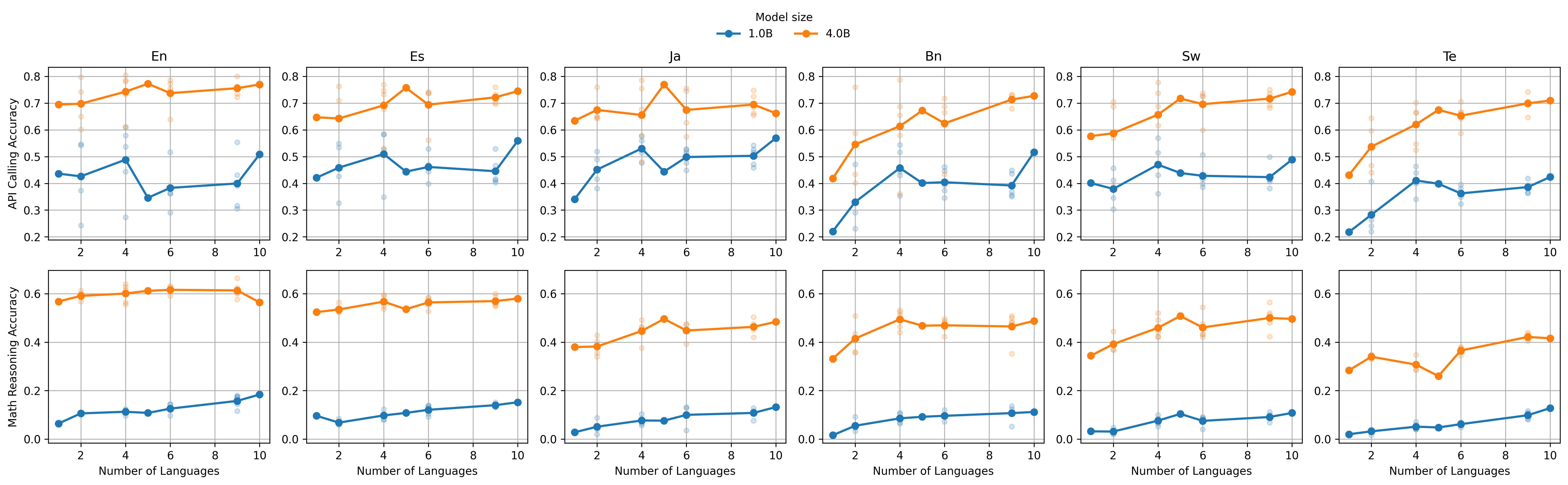}
    \caption{Gemma-3 Models}
    \label{fig:8_s}
\end{subfigure}

\caption{Per-language multilingual scaling trends for (a) Qwen-3 models, and (b) Gemma-3 models. Each cell shows mean accuracy as a function of the number of training languages for API calling (top rows) and math reasoning (bottom rows), with scatter points indicating individual results.}
\label{fig:combined_scaling}
\end{figure*}

Figure~\ref{fig:combined_scaling} presents scaling trends for individual evaluation languages as a function of the number of training languages, for (a) Qwen-3 and (b) Gemma-3 models.

These results complement the aggregated trends shown in Figure~\ref{fig:scaling_langs_all}. Consistent with the main findings, we observe that with the exception of the smallest 0.6B model, increasing the number of training languages generally improves or maintains performance across most languages and model sizes. Gains are particularly pronounced for low-resource languages, while high-resource languages tend to plateau as additional languages are introduced.



\FloatBarrier
\section{Additional Results on Linguistic Diversity Driven Cross-Lingual Generalization}
\label{sec:supp_4_3}

\begin{figure}[h]
    \centering
    \includegraphics[width=\linewidth]{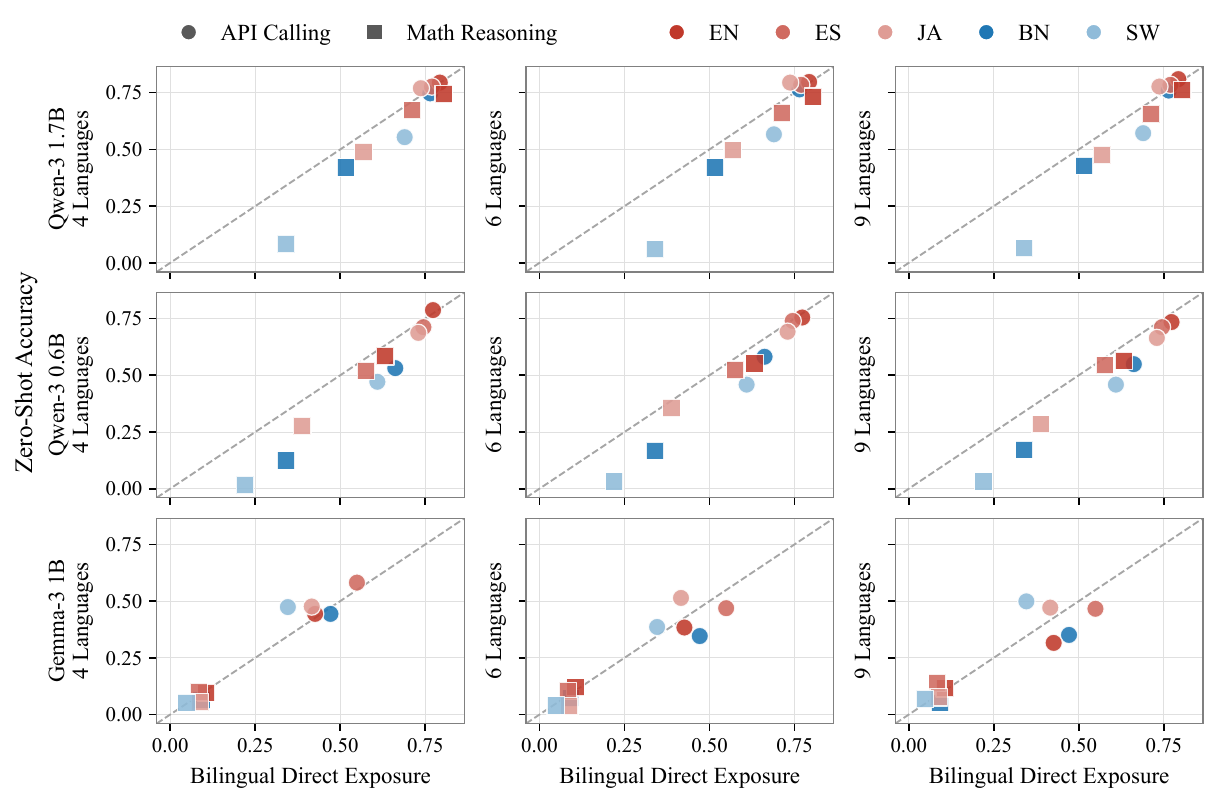}
    \caption{\textbf{Comparison of zero-shot cross-lingual transfer and direct bilingual exposure for smaller models.} Rows correspond to Qwen-3 1.7B (top), Qwen-3 0.6B (middle), and Gemma-3 1B (bottom), with columns showing 4 (left), 6 (middle), and 9 (right) training languages.}
    \label{fig:other_models}
\end{figure}

Figure~\ref{fig:other_models} presents the corresponding plots from Section~4.3 for smaller-scale models (Qwen-3 1.7B, Qwen-3 0.6B, and Gemma-3 1B).

We observe consistent trends with those reported for larger models: high-resource languages (in red) tend to lie close to the diagonal, indicating that language diversity enables strong zero-shot cross-lingual transfer that matches or approaches the performance of direct inclusion at these scales. Low-resource languages often benefit from explicit language inclusion, particularly at the smallest scale (Qwen-3 0.6B). This is consistent with our findings in Section~\ref{sec:result2}, where we observe capacity-driven interference at higher levels of multilinguality for this model.

\end{document}